\documentclass[letterpaper]{article}
\usepackage[margin=1in]{geometry}
\usepackage[T1]{fontenc}
\usepackage{lmodern}
\usepackage{microtype}
\usepackage{xcolor}
\usepackage[hyphens]{url}
\usepackage{graphicx}
\usepackage{natbib}
\usepackage{caption}
\usepackage{algorithm}
\usepackage{algorithmic}
\usepackage{booktabs}
\usepackage{amsmath}
\usepackage{amssymb}
\usepackage{multirow}
\usepackage{float}
\usepackage{placeins}

\newcommand{\hx}[1]{}

\title{Beyond Semantic IDs: Finite-Beam Reachability and Continuous-Query
Generation in Generative Recommendation}
\author{Yu Sun\thanks{Corresponding author} \and Renkai Xiang \and Zuodong Xiang \and Junwei Zhou \and Yike Zhang \and Jianxin Zhang \and Zhihao Zhu \and Xiaoyu Cao \and Hailu Xu}
\date{}

\begin{document}

\maketitle

\begin{abstract}
Generative recommendation (GR) of social media platforms commonly retrieves items by autoregressively decoding discrete Semantic IDs (SIDs) with finite-width
beam search. This work examines how this output interface constrains whether held-out targets
enter finite returned lists, while distinguishing catalog-wide
scoreability from empirical top-$K$ support. We introduce two
continuous-query paradigms: 
% over a shared frozen item space. 
CQG-Single generates one retrieval query in a single pass, and CQG-AR autoregressively
generates continuous residual updates, retaining step-wise
decoding while removing the SID item-token vocabulary, code lookup, and
discrete beam expansion. These designs thereby eliminate the SID-specific reachability constraint induced by finite-beam decoding and avoid the technical challenge of aligning a discrete SID vocabulary with the pretrained backbones' vector space. Both methods make every encoded item directly scoreable
through dense or approximate-nearest-neighbor retrieval. Under a unified
protocol with valid-prefix constrained decoding, 
% target
% support at $K=100$ is stable across seven ML-1M runs
% ($66.67\%\pm0.18\%$ sample SD), although the canonical tokenizer assigns
% unique codes to $99.4\%$ of items. 
CQG-Single exceeds the TIGER-style
baseline in all seven runs 
% (R@10 0.3104 versus 0.2996; exact paired
% sign-flip $p=.0156$), 
while using 55.3\% as many trainable
parameters. CQG-AR remains comparable on ML-1M and is numerically stronger
on Beauty and Sports datasets. A controlled depth ablation further shows that
additional continuous refinement does not monotonically improve shallow
accuracy. These findings motivate decoupling output representation,
autoregressive modeling, and retrieval budget when evaluating generative
recommendation systems.
\end{abstract}

\section{Introduction}

% In today's popular social media platforms, 
Generative recommendation (GR) retrieves items by generating symbolic identifiers rather than directly scoring the entire catalog \citep{li2024large}. The most influential instance of this paradigm is Semantic-ID (SID) recommendation: items are first quantized into discrete code sequences, and an autoregressive model decodes those sequences conditioned on user attributes and interaction history \citep{rajput2023tiger,zheng2024letter,wang2024eager}. This formulation converts recommendation into a sequence-generation problem, and supports constrained decoding through a trie, ensuring that the generated token sequence corresponds to a valid item.

This marks a substantial architectural shift from conventional sequential recommenders. Models such as SASRec encode user histories, user features, and item representations, then score candidate items discriminatively \citep{kang2018sasrec}. In SID-based generative recommendation, the output interface is instead moved to a language-style generator that is trained to emit the SID sequence. Larger pretrained language models can provide useful generic knowledge and may benefit from scaling behavior as the model size increases \citep{kaplan2020scaling}. However, SID-based GR systems usually require three components: First, items must be compressed into a finite hierarchical identifier space, using vector-quantization methods such as VQ-VAE \citep{vandenoord2017vqvae}, residual quantization as in RQ-VAE \citep{lee2022rqvae}, residual or balanced k-means hierarchical clustering, or industrial generative retrieval designs such as OneRec \citep{deng2025onerec}. Second, the generator needs to be trained either by adapting a pretrained LLM through continued pretraining (CPT) and supervised fine-tuning (SFT), where SID tokens become a new vocabulary that must be aligned with the pretrained model's semantic space, or by training a smaller decoder-only model from scratch, which sacrifices pretrained generalization for better SID prediction precision. Third, for inference, this generative model decodes SID sequences from user features and sequential history, then maps the generated codes to item IDs or expands them through item-to-item retrieval. 
% (Figure~\ref{fig:pipeline})
% \hx{should intro Fig 1 here or in the method?}

These steps introduce two practical bottlenecks. 1. Coverage and updatability: a catalog item is useful only if its SID can be generated and mapped back to the item, but newly created items may cause semantic drift between the static clustering model labeling and the item meaning. 2. Support imbalance: a finite hierarchical SID space is a compromise for representing an open-ended catalog. Items are not distributed uniformly across SID leaves, longtail items have worse reachability, and the generated SID set may fail to recall enough valid items even when the related item catalog itself is large.

Because SID inference explores only a prefix-limited subset of the code tree,
we define a \emph{checkpoint- and budget-specific finite-beam reachability
ceiling} as the target-support upper bound induced by a fixed trained model and finite beam budget, even when
the held-out item has a valid code.

% This support limitation motivates a checkpoint- and popularity-dependent
% \emph{crossover pattern}. At shallow retrieval depth, SID models can perform
% well because autoregressive decoding concentrates probability mass on
% high-confidence paths. On rare ML-1M targets for the evaluated seed-42
% checkpoints, TIGER outperforms a continuous alternative at R@10; at some
% deeper $K$, the continuous model catches up or overtakes. We treat this as an
% operating-point-dependent empirical pattern, not as proof that a per-user hit
% rate equals an item-level catalog-coverage ceiling.
This support limitation motivates examining seed-stable gains and \emph{Popularity-Dependent
Crossover}. CQG-Single achieves a stable
aggregate R@10 gain across seven training seeds, and it is
numerically stronger than TIGER-style SID in 38 of 48 popularity--depth
comparisons. 
We also find that TIGER and CQG exhibit different advantages across rare versus popular items and at shallow retrieval depths. Rather than claiming a universal pattern, we characterize the crossover as dependent on the checkpoint, item popularity, and retrieval depth. 
% Overall, CQG performs better across most popularity strata and retrieval depths, while TIGER exhibits only localized reversals at selected points in Q1 and Q4.
% , including all comparisons in Q2--Q3; localized reversals are
% confined to Q1 and Q4, where SID decoding remains competitive at selected
% depths. Thus, 
% Crossover is a localized pattern rather than the dominant
% result. Per-user target support remains distinct from item-level catalog
% coverage.

To isolate this mechanism, we study two complementary continuous-query
formulations. \textbf{CQG-Single} is a deliberately minimal vocabulary-free
model that maps user context to one continuous query in a frozen item space.
\textbf{CQG-AR} retains an autoregressive decoder but replaces
SID-token prediction with continuous residual generation.
% , feeding each generated residual back to refine the query. 
Both retrieve by dense embedding scoring, and approximate nearest-neighbor (ANN) search without a code-to-item lookup.
CQG-Single tests the simplest continuous interface, while CQG-AR controls for
autoregressive model structure.

Our contributions are:
\begin{itemize}
    \item \textbf{Finite-search diagnostics.} For a fixed checkpoint and search
    budget, we distinguish unique-code assignment, target-conditioned beam
    support, catalog-wide scoreability, and empirical union-of-top-$K$
    coverage, and stratify the output-based quantities by item popularity.
    \item \textbf{Conditional crossover pattern.} In some popularity strata
    and evaluated checkpoints, SID decoding can achieve higher shallow
    Recall@$K$, while continuous retrieval can overtake at deeper $K$ under a
    matched returned-list size.
    \item \textbf{Parallel continuous-query methods.} We introduce CQG-Single, a one-step continuous retriever, and CQG-AR, an autoregressive continuous decoder. Their comparison separates the effect of autoregression from the effect of a discrete SID vocabulary.
    \item \textbf{Controlled empirical study.} Across ML-1M, Amazon Beauty,
    and Amazon Sports, we analyze retrieval quality, finite-output coverage,
    objective controls, and LLM-backbone diagnostics while limiting
    scalability claims to the tested catalog sizes.
\end{itemize}

\section{Related Work}

\textbf{Semantic-ID generative recommendation.}
TIGER \citep{rajput2023tiger} introduced hierarchical Semantic IDs using
residual vector quantization and autoregressive decoding. Later work improves
identifier learning and alignment through collaborative semantics, textual
IDs, differentiable indexing, or soft identifiers
\citep{zheng2024letter,wang2024eager,tan2024idgenrec,fu2026diger,li2026unigrec}.
These methods differ in code construction but retain a structured output
interface that typically uses constrained autoregressive decoding. We study
how finite-width search allocates support over that interface, rather than
proposing another discrete tokenizer. This distinction also separates
successful code assignment from successful code retrieval at inference time.

\textbf{Finite search support and popularity.}
Autoregressive tree generation can impose correlation constraints on
representable item distributions \citep{hou2026expressiveness}, while
training-side studies note that tail tokens receive fewer updates. We examine
the complementary inference-side question: even with a valid, nearly unique
code, does a target enter a finite beam? We measure target-conditioned beam
support, returned-list-size-matched catalog coverage, and popularity-stratified behavior,
carefully separating these quantities. In particular, a high unique-code rate
does not imply that every target enters a user's beam, and a target-conditioned
hit rate does not equal catalog coverage. Our experiments treat these as
distinct diagnostics of a trained model and its search budget.

\textbf{Vocabulary-free and continuous recommendation.}
DiffuRec, DreamRec, and ContRec use diffusion or continuous latent generation
\citep{10.1145/3631116,yang2023dreamrec,qu2026contrec}, while GPT4Rec generates
textual retrieval queries \citep{li2023gpt4recgenerativeframeworkpersonalized}. CQG-Single provides a minimal
continuous-query control, and CQG-AR adds step-wise residual refinement. Their
purpose is to separate autoregression from discrete SID output, not to claim
dense retrieval itself as novel. This paired design asks whether step-wise
generation requires discrete identifier tokens: CQG-Single removes both the
token vocabulary and serial decoding, whereas CQG-AR preserves serial
generation but replaces tokens with continuous residual updates.

\paragraph{Adjacent efficient and multimodal model applications.}
Resource-efficient language modeling and multimodal adaptation have been investigated in related application settings. Energy-efficient retrieval-augmented generation with small language models addresses on-device constraints~\citep{cheng2026toward}. In short-video governance, in-prompt process supervision~\citep{liu-etal-2026-ips} and reasoning-enhanced domain-adaptive multimodal pretraining~\citep{wang-etal-2025-reasoning-enhanced} study different supervision and adaptation strategies. Audio-enhanced vision--language modeling with latent-space broadening explores multimodal representation and data expansion~\citep{10.1145/3711896.3737195}. These studies provide broader context, but do not establish results on finite-beam Semantic-ID reachability or the continuous-query recommendation methods analyzed here.

\textbf{Sequential, dense, and hybrid retrieval.}
GRU4Rec, SASRec, and BERT4Rec score catalog items from interaction histories
\citep{hidasi2016gru4rec,kang2018sasrec,sun2019bert4rec}; dual encoders are
established in retrieval and recommendation
\citep{huang2013dssm,covington2016youtube}. Generative retrieval instead
decodes identifiers \citep{tay2022dsi,wang2022nci}, and LIGER combines dense
and generative retrieval \citep{yang2024liger}. These lines establish both
catalog scoring and identifier decoding as known retrieval interfaces.
Our contribution is therefore a controlled comparison in a shared frozen item
space, using CQG-Single and CQG-AR to distinguish finite SID output support
from autoregressive modeling without attributing the comparison to a larger
item encoder or an additional reranker.
\section{Method}

\begin{figure*}[t]
\centering
\includegraphics[width=0.9\textwidth]{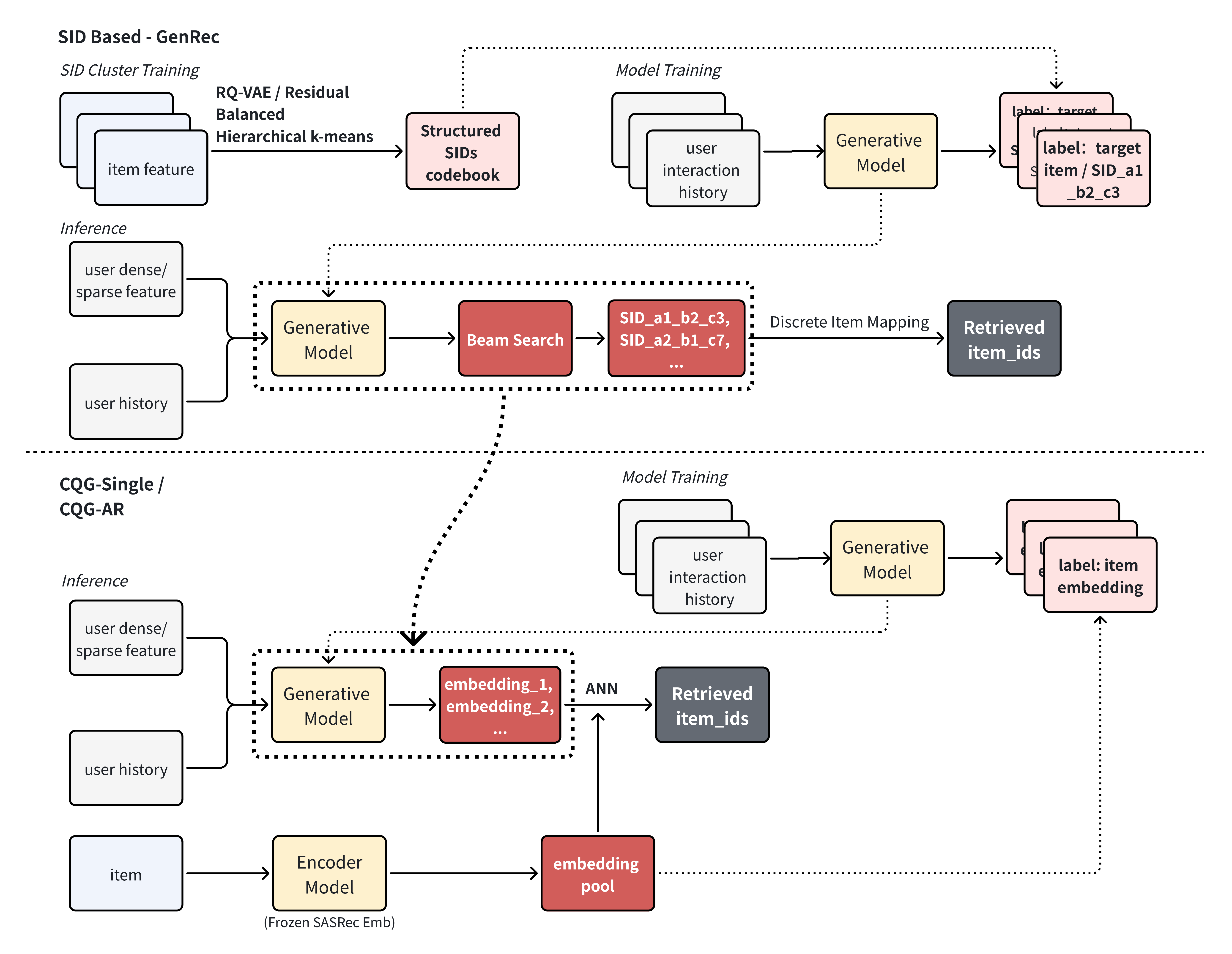}
\caption{Training and inference pipelines for SID retrieval, CQG-Single, and CQG-AR. SID-based GR is trained to predict clustered code sequences, whereas CQG
generates continuous queries in item-embedding space.
SID retrieval autoregressively generates discrete code tokens and explores
catalog-valid paths with finite-prefix beam search. CQG-Single produces one
continuous query in a single forward pass, whereas CQG-AR autoregressively 
generates residual updates that refine a continuous query. Both CQG variants 
retrieve from the same frozen item embedding index by dense scoring or ANN, 
without an item-token vocabulary or code-to-item lookup.}
\label{fig:pipeline}
\end{figure*}

\subsection{Problem Formulation}

Let $\mathcal{I}$ be a catalog of $N$ items and let user $u$ have a chronological interaction sequence $s_u=(i_1,\ldots,i_t)$. The task is to retrieve the next item $i_{t+1}$. SID-based GR maps each item $i$ to a discrete code sequence $z_i=(z_{i,1},\ldots,z_{i,L})$ and trains an autoregressive model $p(z_i\mid s_u)$. Inference uses beam search over valid code prefixes.

Both of our continuous methods assume a frozen item embedding matrix $E\in\mathbb{R}^{N\times d}$, where each row $e_i$ is an L2-normalized item vector. They condition on the same user context $c_u$ available to the SID decoder, including the sequential item history and, when available, demographic features. They differ in whether the continuous output is produced in one step or autoregressively.

\subsection{CQG-Single: Single-Step Continuous Query Generation}

CQG-Single learns a single-step query generator $f_\theta$:
\begin{equation}
q_u = f_\theta(c_u), \quad q_u\in\mathbb{R}^d .
\end{equation}
The recommendation score is cosine similarity:
\begin{equation}
\operatorname{score}(u,i)=q_u^\top e_i .
\end{equation}
The top-K items are retrieved by exact dense scoring or ANN search. Every item in $E$ is scoreable by construction, although catalog-wide scoreability does not guarantee that every item will empirically appear in a finite top-$K$ list.

The query generator is a causal Transformer that maps the final valid history
state back into the shared item space. Unlike SID decoding, it produces the
retrieval query in one forward pass without an item-token vocabulary.

\subsection{CQG-AR: Autoregressive Continuous Query Generation}

CQG-AR retains an autoregressive decoder but replaces discrete SID-vocabulary prediction with continuous residual generation. At decoding step $\ell$, the decoder conditions on the user context and all previous continuous outputs:
\begin{align}
h_\ell &= D_\phi(c_u,\Delta_{<\ell}),\\
\Delta_\ell &= W_{\mathrm{AR}}h_\ell,\\
\hat e_u^{(\ell)} &=
\operatorname{norm}\!\left(\hat e_u^{(\ell-1)}+\Delta_\ell\right).
\end{align}
We initialize the cumulative query as
\(\hat e_u^{(0)}=\mathbf{0}\).
The generated residual $\Delta_\ell$ is projected back into the decoder input space and appended as the next autoregressive input. After $L$ steps, the refined query $\hat e_u^{(L)}$ retrieves items from the same frozen embedding matrix $E$. Thus CQG-AR preserves the conditional, step-wise decoding structure of SID retrieval while removing the SID vocabulary, codebook lookup, and discrete beam expansion.
% We use $L=3$ as an architecture-matched control for the three-level SID output, rather than assuming that three continuous refinements are intrinsically necessary. The $L=1$ setting is its single-step boundary and provides a controlled link to CQG-Single.

\subsection{Training Objectives}

We evaluate four CQG objectives: full-catalog CE, sampled CE, in-batch
InfoNCE, and MSE. For CQG-Single, $q_b$ below is the single generated query;
for CQG-AR, it is the final refinement $\hat e_b^{(L)}$. In the primary
CQG-AR configuration, only the final query receives direct CE supervision;
intermediate supervision is used only in the decoding-depth ablation.

\textbf{Full-catalog cross-entropy.}
Our primary objective treats retrieval as classification over all items:
\begin{equation}
\mathcal{L}_{\mathrm{CE}}
= -\frac{1}{B}\sum_{b=1}^{B}
\log
\frac{\exp(q_b^\top e_b^+/\tau)}
{\sum_{j=1}^{N}\exp(q_b^\top e_j/\tau)} ,
\end{equation}
where $\tau=0.07$. CE gives complete catalog visibility at training time but costs $O(N)$ per step.

\textbf{Sampled cross-entropy.}
To control training-time catalog visibility, sampled CE replaces the full
denominator with the positive item and a sampled set
\(\mathcal S_b\) of negatives:
\begin{equation}
\mathcal{L}_{\mathrm{sCE}}
=-\frac{1}{B}\sum_{b=1}^{B}\log
\frac{\exp(q_b^\top e_b^+/\tau)}
{\exp(q_b^\top e_b^+/\tau)+
\sum_{j\in\mathcal S_b}\exp(q_b^\top e_j/\tau)} .
\end{equation}
The reported control uses 256 sampled negatives.

\textbf{In-batch InfoNCE.}
InfoNCE treats the other target items in the minibatch as negatives:
\begin{equation}
\mathcal{L}_{\mathrm{NCE}}
=-\frac{1}{B}\sum_{b=1}^{B}\log
\frac{\exp(q_b^\top e_b^+/\tau)}
{\sum_{j\in\mathcal B}\exp(q_b^\top e_j^+/\tau)} .
\end{equation}

\textbf{Mean-squared error.}
We also test direct embedding regression:
\begin{equation}
\mathcal{L}_{\mathrm{MSE}}
=\frac{1}{B}\sum_{b=1}^{B}\|q_b-e_b^+\|_2^2 .
\end{equation}
MSE preserves geometric alignment but yields lower retrieval accuracy than CE
in our experiments. BPR is used to pretrain the shared SASRec item encoder,
not as one of the reported CQG objectives.

\subsection{Retrieval and Contrast with SID}

Standalone CQG-Single ranks items according to $q_u^\top e_i$, while CQG-AR uses $(\hat e_u^{(L)})^\top e_i$. Exact catalog scoring can be replaced by ANN search, such as FAISS, for large catalogs. Our CQG-AR experiments retrieve only with the final query \(\hat e_u^{(L)}\); merging candidates from intermediate or interest-specific queries is an untested extension. In contrast, SID retrieval requires codebook learning, autoregressive discrete decoding, beam search, and code-to-item lookup; a finite beam explores only a restricted subset of valid code paths. CQG-Single instead generates one continuous query, while CQG-AR preserves step-wise autoregressive generation with continuous outputs. Both directly query the item embedding index without introducing an item-token vocabulary. (Figure~\ref{fig:pipeline})

% \subsection{Contrast with SID Retrieval}

% Figure~\ref{fig:pipeline} compares the inference mechanisms. SID retrieval requires codebook learning, autoregressive discrete decoding, beam search, and code-to-item lookup. Under a finite beam budget, only a restricted subset of valid code paths is explored. CQG-Single uses single-step continuous generation, whereas CQG-AR retains autoregressive decoding with continuous outputs. Both continuous variants query a dense index without introducing an item-token vocabulary.

\section{Experiments}

\subsection{Experimental Setup}

\textbf{Datasets.}
We use MovieLens-1M (ML-1M), Amazon Beauty, and Amazon Sports. ML-1M is the primary controlled diagnostic because its catalog is small enough for exact full-catalog scoring. Beauty and Sports test cross-domain behavior. Our processed Sports split (21,519 users and 10,693 non-padding items) differs from the commonly reported 5-core split used by TIGER, LETTER, and EAGER; consequently, we do not directly compare our Sports values with their published numbers. (We will use “TIGER” to denote our hierarchical implementation of \(k\) means in TIGER-style.) 
% We exclude Yelp because its SID and continuous baselines are not aligned under one evaluation protocol.

\begin{table}[t]
\centering
\small
\begin{tabular}{lrrrr}
\toprule
Dataset & Users & Items & Interactions & Density \\
\midrule
ML-1M & 6,040 & 3,416 & \textbf{999K} & \textbf{4.84\%} \\
Beauty & \textbf{22,363} & \textbf{12,101} & 198K & 0.07\% \\
Sports & 21,519 & 10,693 & \(\sim\)178K & 0.08\% \\
\bottomrule
\end{tabular}
\caption{Dataset statistics. Item counts exclude the padding row. Density is
defined as $|\mathcal{R}|/(|\mathcal{U}||\mathcal{I}|)$, where
$\mathcal{R}$ is the set of observed interactions, $\mathcal{U}$ the users,
and $\mathcal{I}$ the non-padding items. 
Higher density indicates that users interact with a larger fraction of the
catalog, whereas lower density indicates sparser interactions and a more
pronounced long tail. Bold marks the largest value in each numerical column.}

% The processed Sports split differs
% from the standard 5-core split used in several published SID studies, so its
% results are treated as within-paper diagnostics.}
\label{tab:data}
\end{table}

\begin{figure}[t]
\centering
\includegraphics[width=0.9\columnwidth]{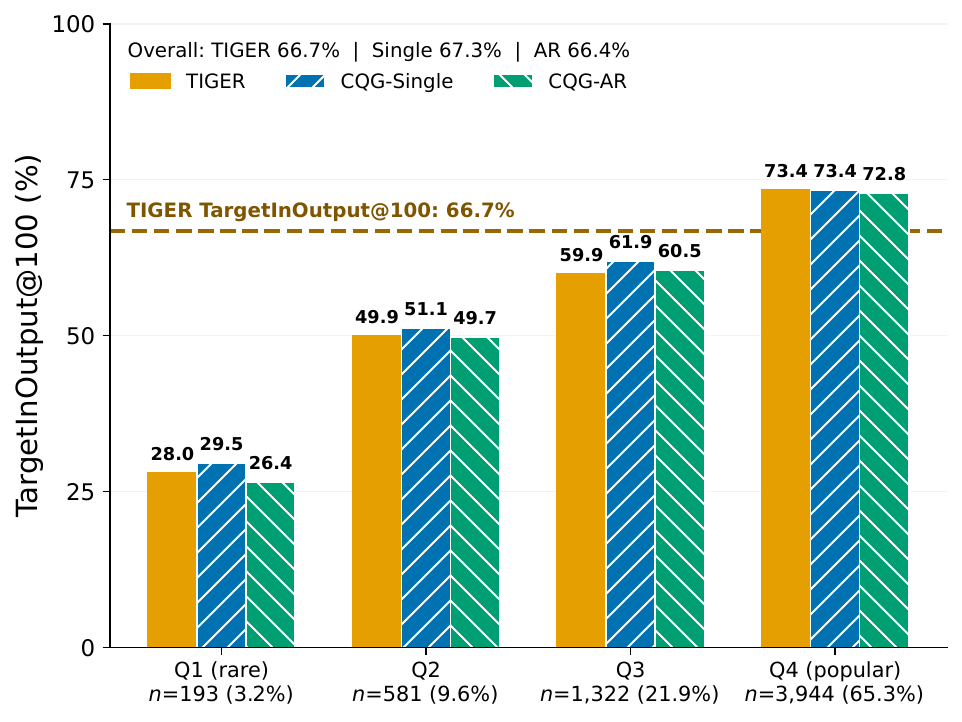}
\caption{ML-1M TargetInOutput@100 by target-popularity quartile (seed 42).
Finite returned lists exhibit strongly popularity-skewed target support,
while CQG-Single matches or exceeds TIGER across all popularity quartiles
at the reported precision.
All methods produce matched Top-200 lists; TargetInOutput@100 is evaluated on their first 100 positions; TIGER uses
valid-prefix constrained decoding. Bars are user weighted, whereas
Table~\ref{tab:target-support-comparison} reports item-balanced quartile values.}

\label{fig:reachability}
\end{figure}

\textbf{Protocol.}
We use leave-one-out evaluation: the last item is held out for testing, the second-to-last item for validation, and earlier interactions for training. Historical items are excluded from ranking. We report Recall@K and NDCG@K, primarily for $K\in\{10,30,50,70,90\}$ in reachability analysis and $K\in\{5,10,20\}$ in cross-dataset evaluation.

\begin{figure*}[t]
\centering
\includegraphics[width=\textwidth]{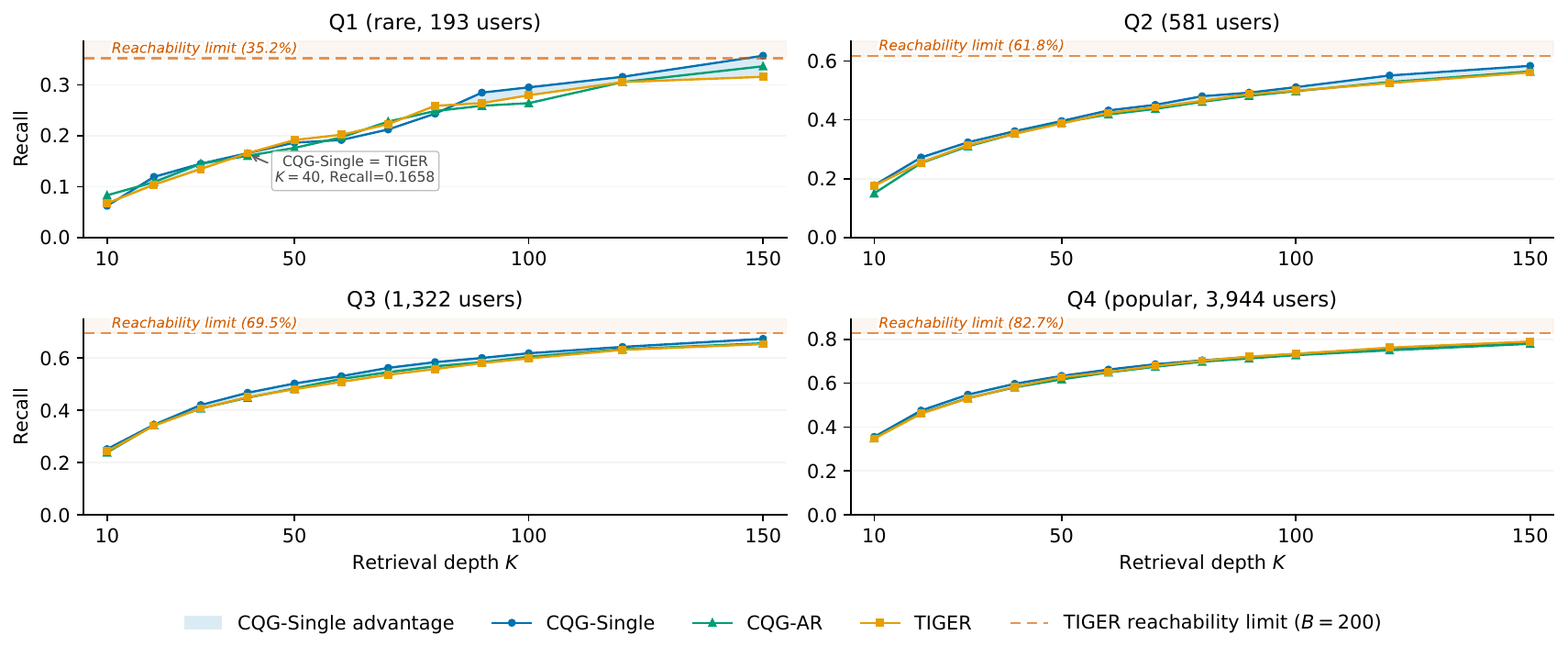}
\caption{Measured per-quartile Recall@$K$ curves on ML-1M, seed 42.
% CQG-Single and CQG-AR are shown separately; 
% Blue shading highlights intervals in which CQG-Single exceeds TIGER. 
Dashed lines show finite-output target-support ceiling of 
TIGER's user-weighted TargetSupport@200 within each quartile.
% , which upper-bounds
% Recall@$K$ for TIGER. 
% the regions above dashline are unreachable for the evaluated model--beam pair. 
% The Q1 annotation marks the exact measured
% CQG-Single--TIGER tie at $K=40$ (Recall $=0.1658$). 
% Every curve point uses saved per-user output at
% $K\in\{10,20,\ldots,100,120,150\}$
% % ; no metric point is interpolated. 
All three methods return exactly 200 valid, distinct, history-filtered items per user. The X-axis represents retrieval depth $K$, the number of
top-K items generated, while the Y-axis reports the fraction of users whose held-out target appears within the Top-$K$ list. CQG-Single outperforms TIGER in 38 of the 48 popularity--depth comparisons, particularly on Q2--Q3. TIGER remains stronger for some rare
Q1 targets at shallow depths.
% , while the two methods are comparable at several depths on Q4. 
These results indicate that CQG's advantage is affected by both target-item popularity and retrieval depth.
This diagnostic uses a separately selected CQG-Single checkpoint
(aggregate R@10 \(=0.3063\)); the primary seed-42 checkpoint in Table 3 has
R@10 \(=0.3036\). Full values for all 12 depths and paired-bootstrap
intervals appear in Appendix C.1.}
\label{fig:crossover}
\end{figure*}

\begin{table}[!t]
\centering
\small
\setlength{\tabcolsep}{4pt}
\renewcommand{\arraystretch}{0.87}
\begin{tabular}{lrrrrrr}
\toprule
& \shortstack{User-weighted\\TargetInOutput@100} &
\multicolumn{4}{c}{Item-balanced TargetInOutput@100} &
\shortstack{TargetItemSupport\\@200} \\
\cmidrule(lr){3-6}
Method & Overall & Q1 rare & Q2 & Q3 & Q4 popular & Distinct targets \\
\midrule
TIGER & 66.7\% & 25.8\% & 45.2\% & 54.0\% & 68.6\% & 78.1\% \\
CQG-Single & \textbf{67.3\%} & \textbf{26.4\%} & \textbf{46.9\%} & \textbf{56.3\%} & \textbf{70.0\%} & \textbf{78.9\%} \\
CQG-AR & 66.4\% & 24.4\% & 45.4\% & 54.8\% & 69.6\% & 78.8\% \\
\bottomrule
\end{tabular}
\caption{ML-1M finite-output support (seed 42). Overall is user weighted;
Q1--Q4 average distinct target items within training-frequency quartiles, and
the final column is the fraction of distinct targets retrieved at least once
by Top-200. Returned-list sizes and filtering are matched across methods.
Bold denotes the best value in each column.}
\label{tab:target-support-comparison}
\end{table}

\textbf{Baselines.} We compare against SASRec \citep{kang2018sasrec}, TIGER-style SID retrieval, and reported SID results from TIGER, LETTER, and EAGER where protocols allow contextual comparison. We also include non-learned dense retrieval baselines: random retrieval, last-item kNN, and mean-of-5 kNN.

\textbf{Model training protocol.}
We first train a two-layer SASRec with BPR on chronological next-item pairs and select its checkpoint by full-catalog validation NDCG@10. Its 64-dimensional item embedding table is then frozen and shared by both continuous methods as the history-input table, target embedding or classifier matrix, and retrieval index. The main experiments use no textual or language-model item encoder.

\textbf{CQG-Single} is a four-layer, four-head causal Transformer with width
256. It directly generates one normalized query; the final configuration uses
full-catalog CE, with MSE and in-batch InfoNCE evaluated as ablations.
\textbf{CQG-AR} uses a four-layer, six-head causal Transformer with width 384
to autoregressively generate three residual query updates. Each residual is
fed into the next decoding step, while the cumulative query is updated by
adding the residual and then normalizing the result. In the final
configuration, only the final query receives direct full-catalog CE
supervision. The multimodal Sports diagnostic instead uses a frozen fusion of
collaborative and ImageNet-pretrained ResNet-18 features. Full implementation
details are provided in Appendix D.

\begin{table}[!t]
\centering
\small
\setlength{\tabcolsep}{3pt}
\renewcommand{\arraystretch}{0.87}
\resizebox{\linewidth}{!}{%
\begin{tabular}{llrrrrr}
\toprule
Method & Config. & Trainable Params & \shortstack{ML-1M\\R@10} & \shortstack{ML-1M\\NDCG@10} & \shortstack{Beauty\\R@10} & \shortstack{Sports\\R@10} \\
\midrule
SASRec & Full-rank & \shortstack{0.32 / 0.88 / 0.79M} & 0.2797 & 0.1535 & 0.0558 & \textbf{0.0390} \\
TIGER & Discrete AR & 5.78M & 0.2996 \(\pm\) 0.0072 & 0.1739 \(\pm\) 0.0046 & 0.0477 \(\pm\) 0.0042 & 0.0236 \(\pm\) 0.0028 \\
CQG-Single & One-step & 3.2M & \textbf{0.3104 \(\pm\) 0.0047} & \textbf{0.1822 \(\pm\) 0.0044} & \textbf{0.0605 \(\pm\) 0.0046} & 0.0363 \(\pm\) 0.0015 \\
CQG-AR & AR continuous (\(L=3\)) & 5.62M & 0.2976 \(\pm\) 0.0032 & 0.1735 \(\pm\) 0.0016 & 0.0571 \(\pm\) 0.0043 & 0.0355 \(\pm\) 0.0004 \\
\bottomrule
\end{tabular}%
}
% \caption{Cross-dataset primary comparison. SASRec entries are single-checkpoint
% full-rank baselines; TIGER, CQG-Single, and CQG-AR report mean $\pm$ sample SD.
% ML-1M uses seven training seeds (42--48), whereas Beauty and Sports use three
% (42--44); seed-level values and 95\% $t$ intervals are in
% Appendix B. SASRec parameter counts are dataset dependent because its
% item embedding table scales with catalog size. Each ML-1M TIGER checkpoint is
% reselected from the saved 10K--50K-step checkpoints using valid-prefix
% constrained validation; the corrected R@10 mean is unchanged from the earlier
% selection, while deeper returned-list metrics change. All datasets use aligned
% full-user evaluation within each dataset. Amazon TIGER uses our
% hierarchical-$k$-means baseline with constrained-validation checkpoint
% selection and valid-prefix decoding.}
\caption{Cross-dataset primary comparison. TIGER and CQG results are
mean $\pm$ sample SD over seeds 42--48 on ML-1M and seeds 42--44 on
Beauty and Sports; SASRec uses one full-rank checkpoint.  SASRec's trainable
parameter counts of 0.32/0.88/0.79M correspond to ML-1M, Beauty, and
Sports, respectively, reflecting their different item-embedding table
sizes. All methods use aligned full-user evaluation, and TIGER uses
valid-prefix constrained decoding. Seed-level results and 95\% $t$
intervals are reported in Appendix B. Bold denotes the best value in each
dataset/metric column.}
\label{tab:primary-comparison}
\end{table}

\section{Results}

\subsection{Budget-Specific Finite-Beam Reachability}

For a fixed model and returned-list size $B$, target support is the fraction of test
users whose held-out target occurs anywhere in the returned list. On ML-1M,
the seed-42 TargetInOutput@100 values are 66.7\% for TIGER, 67.3\% for
CQG-Single, and 66.4\% for CQG-AR. This is a per-user quantity closely related
to Recall@100.
% ; it is not the fraction of catalog items retrieved.
Figure~\ref{fig:reachability} reports the matched popularity breakdown and
user counts.

For this fixed model--search pair, TargetSupport@$B$ is a realized ceiling on
the Recall of any downstream reranker restricted to the decoded beam: a target
that never enters the candidate set cannot be recovered later. This is a
finite-beam reachability ceiling, not an architecture-invariant claim; increasing
$B$ can raise it. CQG avoids this particular beam-exclusion mechanism because
every encoded item remains scoreable, although its finite Top-$K$ Recall is
still determined by query quality.

Table~\ref{tab:target-support-comparison} provides the user-weighted and returned-list-size-matched finite-output comparison. TargetInOutput@100 is user weighted. Q1--Q4 first compute the hit rate of each distinct test-target item and then average items equally within training-frequency quartiles. TargetItemSupport@200 is the fraction of distinct held-out target items retrieved for at least one associated test instance. 
% Under the corrected constrained protocol, 
CQG-Single retains advantage in overall user-weighted TargetInOutput@100 and in all four item-balanced quartiles.
% whereas TIGER slightly exceeds CQG-AR on rare targets.

Union-of-output catalog coverage is reported separately in Table~\ref{tab:matched-coverage}. On Beauty, and more modestly at Top-100 on ML-1M, CQG achieves broader
catalog coverage than TIGER, suggesting potential for improved retrieval
and generalization to long-tail items.
% Figure~\ref{fig:reachability} is user weighted, whereas the Q1--Q4 columns in
% Table~\ref{tab:target-support-comparison} average distinct target items equally
% within the same quartile mapping.
This result separates code assignment from target support: 99.4\% unique codes do not imply that every held-out target enters a finite returned list. Under the corrected Top-100 protocol, 33.3\% of held-out user targets do not enter the corresponding candidate list; this is not a statement that 33.3\% of catalog items are never retrieved. The distinct union-of-output catalog-coverage diagnostic is reported in Table~\ref{tab:matched-coverage}.

\subsection{Seed-Stable Gains and Popularity-Dependent Crossover}

Table~\ref{tab:primary-comparison} shows that CQG-Single outperforms
TIGER-style SID in all seven ML-1M training runs, with a mean paired R@10
gain of 0.0107 (95\% paired $t$ interval [0.0021, 0.0193]; exact sign-flip
$p=.0156$). For the separately selected seed-42 diagnostic checkpoint,
Figure~\ref{fig:crossover} shows numerically higher
CQG-Single Recall in 38 of the 48 popularity--depth comparisons, including
all 24 comparisons in Q2--Q3; the reversals are confined to Q1 and Q4.
Full curve values and paired intervals are reported in Appendix C.1. The
curve is checkpoint-specific and is not substituted for the primary
seed-42 value in Table 3. Together, these results show that CQG-Single's
ML-1M gain is stable across training seeds and extends across most evaluated
popularity--depth cells, without implying universal dominance.

CQG-AR retains step-wise autoregressive decoding while replacing discrete
SID tokens with continuous residual outputs. It remains comparable to
TIGER-style SID on ML-1M and achieves higher mean R@10 on both Amazon
datasets. The largest gain occurs on Sports, where R@10 increases from
0.0236 to 0.0355 (50.4\% relative); the seed-42 paired difference is 0.0146
(95\% bootstrap CI [0.0119, 0.0173], $p<.0002$). Thus, continuous outputs
can potentially remove SID-specific decoding constraints without sacrificing
autoregressive generalization or retrieval accuracy.

\section{Analysis}

\paragraph{Effect of Autoregressive Refinement.}
Under the controlled depth-ablation recipe with intermediate-loss weight 0.3,
increasing CQG-AR depth from one to four steps does not monotonically improve
shallow retrieval. R@10 varies only from 0.2909 to 0.2944, and the single-step
boundary ($L=1$) is competitive with the three-step control ($L=3$).
The final primary CQG-AR recipe instead uses intermediate-loss weight 0, so
this experiment isolates depth only within the ablation recipe. Additional refinement is more visible at the deeper cutoff:
R@100 increases from 0.6550 at $L=1$ to 0.6697 at $L=4$, although the trend is
not monotonic at every intermediate depth. These results suggest that serial
continuous refinement may adjust deeper candidate ordering without reliably
improving shallow accuracy. We therefore use $L=3$ primarily to match the
three-level SID decoding structure, rather than because three continuous
outputs are intrinsically necessary.

\paragraph{Objective visibility.}
Full-catalog CE gives continuous queries access to every item at each update,
whereas SID decoding receives token-level supervision. To control this
difference, we train CQG-AR with 256 sampled negatives. Sampled CE reduces
R@10 from 0.2982 to 0.2895 while retaining performance close to the
TIGER-style baseline. Thus training-time catalog visibility explains part,
but not all, of the observed behavior; our architecture-level claim is
limited to eliminating discrete code lookup and finite-beam exclusion.
The sampled-negative control and broader frozen-item objective comparison are
reported in Appendix F.1.

% \paragraph{Scoreability versus empirical output coverage.}
% Catalog-wide scoreability does not imply that every item appears in a finite
% output set. With a matched returned-list size of 10 on ML-1M, TIGER, CQG-Single, and
% CQG-AR cover 70.1\%, 69.7\%, and 68.7\% of the catalog, respectively. Thus,
% CQG-Single's higher mean R@10 is not explained by retrieving a larger union of
% distinct items at this cutoff. At Top-100, the two continuous methods cover
% 97.6\% of the catalog, compared with 97.0\% for TIGER. The separation is
% larger on Beauty at Top- (74.4\%, 89.5\%, and 87.9\%), although all methods
% approach full union coverage by Top-100. These results reinforce the
% distinction among scoreability, per-user target support, and empirical catalog
% coverage. Sports provides a counterexample to equating coverage with accuracy:
% TIGER has the broadest union coverage at both cutoffs, while full-rank SASRec
% has the highest R@10.

\paragraph{Robustness across SID constructions.}
To examine whether the popularity skew is specific to our
hierarchical-$k$-means baseline, we additionally evaluate a closer
T5/RQ-VAE-style reproduction on Beauty. Across decoder seeds 42--44,
99.48\% of generated sequences correspond to valid SIDs, yet the union of
the Top-100 outputs covers only \(44.0\%\pm13.7\%\) of the catalog.
Coverage increases sharply from \(14.0\%\pm7.6\%\) for Q1 rare items to
\(88.7\%\pm13.0\%\) for Q4 popular items. This qualitative gradient suggests
that popularity-skewed finite-output coverage is not unique to our
hierarchical-$k$-means SID construction. Because this reproduction remains
below the reference implementation in ranking accuracy, we treat it as a
robustness diagnostic rather than a reproduced TIGER benchmark. Full
details are reported in Appendix E.2 and
Table~\ref{tab:faithful-coverage}.

\begin{table*}[!t]
\centering
\small
\setlength{\tabcolsep}{2.5pt}
\renewcommand{\arraystretch}{0.87}
\begin{tabular}{lrr|rr|rr}
\toprule
Dataset &
\shortstack{TIGER\\Cov.@10} &
\shortstack{TIGER\\Cov.@100} &
\shortstack{CQG-Single\\Cov.@10} &
\shortstack{CQG-Single\\Cov.@100} &
\shortstack{CQG-AR\\Cov.@10} &
\shortstack{CQG-AR\\Cov.@100} \\
\midrule
ML-1M & \textbf{70.1\%} & 97.0\% & 69.7\% & \textbf{97.6\%} & 68.7\% & \textbf{97.6\%} \\
Beauty & 74.4\% & 96.3\% & \textbf{89.5\%} & \textbf{99.9\%} & 87.9\% & 99.8\% \\
Sports & \textbf{94.0\%} & \textbf{99.5\%} & 63.1\% & 96.9\% & 71.8\% & 98.4\% \\
\bottomrule
\end{tabular}
\caption{Empirical union-of-Top-$K$ catalog coverage under matched
returned-list sizes (seed 42). All methods use the same full-user sets
and history filtering. Bold denotes the highest item coverage within each
dataset and cutoff.}
\label{tab:matched-coverage}
\end{table*}

\begin{table}[!t]
\centering
\small
\setlength{\tabcolsep}{6pt}
\renewcommand{\arraystretch}{1.00}
\begin{tabular}{lrrr}
\toprule
Depth & R@10 & NDCG@10 & R@100 \\
\midrule
\(L=1\) & 0.2924 & \textbf{0.1719} & 0.6550 \\
\(L=2\) & 0.2909 & 0.1688 & 0.6525 \\
\(L=3\) & 0.2914 & 0.1667 & 0.6619 \\
\(L=4\) & \textbf{0.2944} & 0.1717 & \textbf{0.6697} \\
\bottomrule
\end{tabular}
\caption{Single-seed CQG-AR decoding-depth ablation on ML-1M. All rows use
seed 42, full-catalog CE, and intermediate-loss weight 0.3. Within this
controlled ablation recipe, additional continuous refinements do not yield a
monotonic R@10 or NDCG@10 improvement; the final primary recipe instead uses
intermediate-loss weight 0.}
\label{tab:ar-depth}
\end{table}

\begin{table}[!htbp]
\centering
\small
\begin{tabular}{lrr}
\toprule
Catalog group & Coverage@100 & Std. \\
\midrule
Q1 rare & 14.0\% & 7.6\% \\
Q2 & 26.1\% & 13.8\% \\
Q3 & 47.4\% & 20.3\% \\
Q4 popular & \textbf{88.7\%} & 13.0\% \\
\midrule
Overall & 44.0\% & 13.7\% \\
\bottomrule
\end{tabular}
\caption{Empirical union-of-top-100 catalog coverage for the closer T5/RQ-VAE
Beauty reproduction, averaged over decoder seeds 42--44. Bold denotes the
highest quartile coverage.}
\label{tab:faithful-coverage}
\end{table}

\paragraph{Returned-list-size sensitivity.}
For the seed-42 ML-1M TIGER checkpoint selected by constrained validation,
increasing the effective returned-list size from 100 to 200 raises
TargetInOutput from 66.7\% to 76.3\%. Across seven seeds, the corresponding
means are \(0.6667\pm0.0018\) and \(0.7593\pm0.0025\). The evaluator increases
the internal beam when deduplication and history filtering would otherwise
leave too few items. Larger returned lists therefore relax the observed
support limitation; our evidence supports a finite-compute trade-off, not an
architecture-invariant catalog ceiling.

\section{Discussion and Limitations}

\paragraph{Practical implications.}
Continuous-query retrieval removes discrete code lookup and finite-beam
exclusion, while supporting exact or ANN retrieval without changing the
output dimensionality as encoded items are added. The controlled depth
ablation suggests that CQG-Single is preferable when shallow accuracy and
inference efficiency are primary, whereas CQG-AR preserves step-wise
autoregressive conditioning when this structure is desired. However,
full-rank SASRec remains strongest on Sports, showing that continuous
outputs remove a search constraint but do not guarantee superior ranking
accuracy. Production-scale latency, memory, index-update cost, and
cold-item accuracy remain unmeasured.

\paragraph{Interpretation of the crossover.}
The observed crossover is a checkpoint-, popularity-, and
retrieval-depth-dependent empirical pattern
rather than a universal ranking theorem. On ML-1M, TIGER is numerically
stronger for rare targets at some shallow and intermediate cutoffs, whereas
CQG-Single is generally stronger for more popular targets and at several
deeper cutoffs.
% ; none of the aggregate paired ML-1M differences is significant
% at the 0.05 level. 
Beauty does not reproduce the rare-item shallow advantage,
further limiting claims of universality. These results support a conditional
tradeoff between concentrated finite-beam decoding and continuous retrieval,
not universal dominance by either interface. 

% \paragraph{Practical implications.}
% Continuous-query retrieval removes discrete code lookup and finite-beam
% exclusion, and can use exact scoring or ANN retrieval without changing the
% output dimensionality when encoded items are inserted. In the
% protocol-consistent comparison in Table~\ref{tab:primary-comparison},
% CQG-Single has the highest mean shallow accuracy, while CQG-AR shows that
% autoregressive modeling can be retained with continuous outputs. However,
% under the controlled depth-ablation recipe, additional continuous refinement
% does not monotonically improve shallow accuracy, and a single generated query
% remains competitive with three-step refinement. On Sports, full-rank SASRec
% remains strongest, showing that removing discrete output constraints does not
% by itself guarantee superior ranking accuracy. The resulting latency, memory, index-update cost, and cold-item
% accuracy have not been measured at production scale.

\paragraph{Limitations.}
Our evidence is specific to the evaluated model--search--compute
configurations. Wider beams substantially raise TIGER target support, and our
hierarchical-$k$-means TIGER-style baseline is weaker than published SID
systems. The closer T5/RQ-VAE Beauty reproduction also does not match the
reference implementation's ranking accuracy, so its absolute coverage effect
must not be generalized to a fully reproduced TIGER checkpoint. Matched-target
analysis remains observational rather than a causal intervention on
popularity. Full-catalog CE costs
$O(N)$ per update and has only been evaluated on the reported catalog sizes;
billion-item training, ANN latency, memory, and dynamic-catalog accuracy remain
open.

% \section{Conclusion}

% This paper studies finite-beam output support as a practical bottleneck in SID-based generative recommendation. In our evaluated models, target-conditioned support and empirical output coverage can be incomplete and popularity-skewed, and a checkpoint- and popularity-dependent crossover can emerge between SID decoding and vocabulary-free continuous retrieval. CQG-Single demonstrates a minimal one-step continuous query, while CQG-AR shows that an autoregressive decoder can replace discrete SID tokens with continuous residual outputs and retain competitive accuracy. Under the controlled depth-ablation recipe, a single generated query remains competitive with three-step refinement; this result is not asserted as a depth ablation of the final zero-intermediate-loss recipe. Together these results separate the value of autoregression from the constraints of a discrete output vocabulary and motivate protocol-matched evaluation of accuracy, target support, and returned-list-size-matched catalog coverage.

\section{Conclusion}
We study finite-beam output support in SID-based generative recommendation.
Across the evaluated models, target support and catalog coverage can be
incomplete and popularity-skewed, with crossover patterns depending on
checkpoint, popularity, and retrieval depth. CQG-Single generates one
continuous query, whereas CQG-AR replaces SID tokens with autoregressive
residual vectors while retaining competitive accuracy. A controlled CQG-AR
ablation finds no consistent shallow benefit from refinement. CQG-Single is not proposed as the invention of dense retrieval, but a minimal, vocabulary-free interface that enables a controlled comparison with SID under a shared item space.
Together, these results decouple autoregression from discrete output
constraints and motivate matched evaluation of accuracy, target support,
and catalog coverage.

\clearpage
% Integrated supplementary material follows the main text.
\pagestyle{plain}
\onecolumn
\appendix
\setcounter{secnumdepth}{2}

\section{Additional Cross-Dataset Diagnostics}
\label{app:auto-completed}
% BEGIN AUTO_RESULTS
Under the matched Amazon protocol, CQG-Single achieves the best R@10 and
NDCG@10 in all Beauty runs and the best R@10 in two of three Sports runs.
The comparison is against our TIGER-style baseline, not published TIGER.
\begin{table}[H]
\centering
\small
\begin{minipage}[t]{0.48\textwidth}
\centering
\begin{tabular}{llrr}
\toprule
\multicolumn{4}{c}{\textbf{Beauty}} \\
\cmidrule(lr){1-4}
Seed & Method & R@10 & NDCG@10 \\
\midrule
42 & TIGER & 0.0464 & 0.0262 \\
 & CQG-Single & \textbf{0.0552} & \textbf{0.0342} \\
 & CQG-AR & 0.0527 & 0.0321 \\
43 & TIGER & 0.0443 & 0.0261 \\
 & CQG-Single & \textbf{0.0632} & \textbf{0.0389} \\
 & CQG-AR & 0.0612 & 0.0363 \\
44 & TIGER & 0.0525 & 0.0295 \\
 & CQG-Single & \textbf{0.0631} & \textbf{0.0379} \\
 & CQG-AR & 0.0573 & 0.0333 \\
\bottomrule
\end{tabular}
\end{minipage}\hfill
\begin{minipage}[t]{0.48\textwidth}
\centering
\begin{tabular}{llrr}
\toprule
\multicolumn{4}{c}{\textbf{Sports}} \\
\cmidrule(lr){1-4}
Seed & Method & R@10 & NDCG@10 \\
\midrule
42 & TIGER & 0.0205 & 0.0109 \\
 & CQG-Single & 0.0350 & \textbf{0.0192} \\
 & CQG-AR & \textbf{0.0351} & 0.0189 \\
43 & TIGER & 0.0242 & 0.0126 \\
 & CQG-Single & \textbf{0.0361} & \textbf{0.0200} \\
 & CQG-AR & 0.0353 & 0.0189 \\
44 & TIGER & 0.0261 & 0.0135 \\
 & CQG-Single & \textbf{0.0379} & \textbf{0.0215} \\
 & CQG-AR & 0.0359 & 0.0190 \\
\bottomrule
\end{tabular}
\end{minipage}
\caption{Complete Beauty and Sports seed-level results underlying the means
and sample standard deviations in main-paper Table 3. Beauty and Sports use
22,363 and 21,519 aligned test users, respectively. TIGER checkpoints are
selected with constrained validation and evaluated with valid-prefix
decoding; all three methods use the same processed split and history
filtering within each dataset. Bold denotes the best value for each
dataset--seed pair and metric.}
\label{tab:auto-tiger-results}
\end{table}

The 95\% \(t\)-intervals for Beauty R@10 are
\([0.0372,0.0583]\) for TIGER, \([0.0490,0.0720]\) for CQG-Single, and
\([0.0465,0.0676]\) for CQG-AR. For Sports they are
\([0.0165,0.0307]\), \([0.0327,0.0399]\), and
\([0.0344,0.0365]\), respectively. Both continuous intervals on Sports lie
above the TIGER interval; the Beauty intervals overlap. These are
training-run, not user-level, intervals.

\section{Training-Seed Stability}
\label{app:seed-stability}

CQG-Single exceeds TIGER in R@10 in all seven ML-1M runs
(Table~\ref{tab:ml1m-seven-seed}). The mean paired difference is 0.0107
(95\% $t$ interval [0.0021, 0.0193]; exact sign-flip $p=.0156$).
TIGER TargetInOutput is \(0.6667\pm0.0018\) at output size 100 and
\(0.7593\pm0.0025\) at size 200.

All methods share the frozen item table, split, and evaluation history.
Seeds change downstream training; for TIGER they also change hierarchical
$k$-means. ``Paired seed'' therefore means matched replicate index, not
identical random draws across architectures.

\begin{table}[H]
\centering
\small
\setlength{\tabcolsep}{5pt}
\begin{tabular}{lrrrrrr}
\toprule
Seed & \multicolumn{2}{c}{TIGER} & \multicolumn{2}{c}{CQG-Single} &
\multicolumn{2}{c}{CQG-AR} \\
\cmidrule(lr){2-3}\cmidrule(lr){4-5}\cmidrule(lr){6-7}
& R@10 & NDCG@10 & R@10 & NDCG@10 & R@10 & NDCG@10 \\
\midrule
42 & 0.2992 & \textbf{0.1756} & \textbf{0.3036} & 0.1736 & 0.2972 & 0.1720 \\
43 & 0.3038 & 0.1751 & \textbf{0.3161} & \textbf{0.1874} & 0.3028 & 0.1750 \\
44 & 0.2841 & 0.1644 & \textbf{0.3144} & \textbf{0.1831} & 0.2945 & 0.1711 \\
45 & 0.3028 & 0.1768 & \textbf{0.3103} & \textbf{0.1830} & 0.3010 & 0.1755 \\
46 & 0.3045 & 0.1781 & \textbf{0.3139} & \textbf{0.1849} & 0.2969 & 0.1739 \\
47 & 0.2995 & 0.1728 & \textbf{0.3089} & \textbf{0.1832} & 0.2970 & 0.1734 \\
48 & 0.3036 & 0.1747 & \textbf{0.3053} & \textbf{0.1804} & 0.2940 & 0.1738 \\
\midrule
Mean & 0.2996 & 0.1739 & \textbf{0.3104} & \textbf{0.1822} & 0.2976 & 0.1735 \\
Sample SD & 0.0072 & 0.0046 & 0.0047 & 0.0044 & 0.0032 & 0.0016 \\
\bottomrule
\end{tabular}
\caption{Seven-seed ML-1M results under valid-prefix constrained TIGER
decoding. Each TIGER row uses the checkpoint reselected by constrained
validation and evaluates all 6,040 test users with the same history filtering.
All metrics use the standard NDCG@10 notation. Bold denotes the best result
for each seed/metric and for the mean row.}
\label{tab:ml1m-seven-seed}
\end{table}

% END AUTO_RESULTS

\section{Experimental Protocol and Reproducibility}
\label{app:protocol}

\subsection{MSE-trained candidate generation and reranking.}
\label{app:mse-rerank}
The matched pipeline retrieves 100 distinct, history-filtered candidates and
reranks them with the same pretrained SASRec. It evaluates the MSE-trained
retrievers, not the CE-trained primary models.

Across ML-1M, Beauty, and Sports, CQG-Single followed by SASRec reranking
reaches 97.7--102\% of full-rank SASRec Recall@10. CQG-AR nearly recovers
full-rank accuracy on ML-1M and Sports but remains lower on Beauty. The
learned retrievers also outperform random and nearest-neighbor candidate
generation. These are fixed-checkpoint, not multi-seed, results.

\begin{table}[H]
\centering
\small
\begin{minipage}[t]{0.625\textwidth}
\centering
\textbf{(a) Standalone and reranked accuracy}\\[2pt]
\setlength{\tabcolsep}{2.2pt}
\begin{tabular}{lrrr}
\toprule
Method & ML-1M & Beauty & Sports \\
\midrule
\multicolumn{4}{l}{\textit{Recall@10}} \\
SASRec full-rank & .2797 & .0558 & \textbf{.0390} \\
Single MSE standalone & .2425 & .0524 & .0307 \\
Single MSE + SASRec & .2801 & \textbf{.0569} & .0381 \\
AR MSE standalone (s42) & .2414 & .0466 & .0294 \\
AR MSE + SASRec (s42) & .2791 & .0486 & .0378 \\
TIGER + SASRec & \textbf{.2864} & -- & -- \\
\midrule
\multicolumn{4}{l}{\textit{NDCG@10}} \\
SASRec full-rank & .1535 & .0296 & \textbf{.0197} \\
Single MSE standalone & .1424 & .0266 & .0150 \\
Single MSE + SASRec & .1554 & \textbf{.0308} & .0194 \\
AR MSE standalone (s42) & .1366 & .0276 & .0160 \\
AR MSE + SASRec (s42) & .1543 & .0265 & .0194 \\
TIGER + SASRec & \textbf{.1571} & -- & -- \\
\bottomrule
\end{tabular}
\end{minipage}\hfill
\begin{minipage}[t]{0.36\textwidth}
\centering
\textbf{(b) Candidate hierarchy}\\[2pt]
\setlength{\tabcolsep}{2.2pt}
\begin{tabular}{lrr}
\toprule
Generator & ML-1M & Sports \\
\midrule
Random & .0252 & .0037 \\
Last-item kNN & .2349 & .0311 \\
Mean-of-5 kNN & .2545 & .0310 \\
Single + SASRec & \textbf{.2801} & .0381 \\
AR + SASRec & .2791 & .0378 \\
SASRec full-rank & .2797 & \textbf{.0390} \\
\bottomrule
\end{tabular}
\end{minipage}
\caption{Matched Top-100 MSE candidate-generation and reranking diagnostics.
(a) Standalone and SASRec-reranked accuracy; CQG-AR uses the available
seed-42 outputs, and TIGER reranking is available only for ML-1M.
(b) Candidate-generation hierarchy under the same SASRec reranker; only
SASRec full-rank bypasses candidate generation. Bold denotes the best result
in each dataset and metric block.}
\label{tab:mse-rerank}
\label{tab:retrieval-hierarchy}
\end{table}

\subsection{Paired Crossover and Candidate-Generation Diagnostics}
\label{app:crossover-values}

\begin{table}[H]
\centering
\small
\setlength{\tabcolsep}{1.45pt}
\begin{tabular}{llrrrrrrrrrrrr}
\toprule
Quartile & Method & 10 & 20 & 30 & 40 & 50 & 60 & 70 & 80 & 90 & 100 & 120 & 150 \\
\midrule
Q1 rare & Single & .0622 & \textbf{.1192} & \textbf{.1451} & \textbf{.1658} & .1865 & .1917 & .2124 & .2435 & \textbf{.2850} & \textbf{.2953} & \textbf{.3161} & \textbf{.3575} \\
($n=193$) & AR & \textbf{.0829} & .1088 & \textbf{.1451} & .1606 & .1762 & .1969 & \textbf{.2280} & .2487 & .2591 & .2642 & .3057 & .3368 \\
& TIGER & .0674 & .1036 & .1347 & \textbf{.1658} & \textbf{.1917} & \textbf{.2021} & .2228 & \textbf{.2591} & .2642 & .2798 & .3057 & .3161 \\
\midrule
Q2 & Single & \textbf{.1773} & \textbf{.2719} & \textbf{.3236} & \textbf{.3614} & \textbf{.3959} & \textbf{.4320} & \textbf{.4509} & \textbf{.4802} & \textbf{.4923} & \textbf{.5112} & \textbf{.5508} & \textbf{.5835} \\
($n=581$) & AR & .1497 & .2530 & .3098 & .3528 & .3907 & .4182 & .4372 & .4613 & .4819 & .4974 & .5284 & .5645 \\
& TIGER & .1756 & .2547 & .3133 & .3528 & .3873 & .4234 & .4423 & .4647 & .4871 & .4991 & .5250 & .5611 \\
\midrule
Q3 & Single & \textbf{.2519} & \textbf{.3449} & \textbf{.4206} & \textbf{.4675} & \textbf{.5030} & \textbf{.5310} & \textbf{.5628} & \textbf{.5847} & \textbf{.6006} & \textbf{.6188} & \textbf{.6430} & \textbf{.6740} \\
($n=1{,}322$) & AR & .2390 & .3434 & .4077 & .4486 & .4841 & .5204 & .5461 & .5688 & .5840 & .6051 & .6339 & .6566 \\
& TIGER & .2436 & .3411 & .4077 & .4508 & .4811 & .5091 & .5363 & .5582 & .5809 & .5991 & .6316 & .6536 \\
\midrule
Q4 popular & Single & \textbf{.3555} & \textbf{.4754} & \textbf{.5479} & \textbf{.5974} & \textbf{.6331} & \textbf{.6620} & \textbf{.6864} & \textbf{.7039} & .7186 & \textbf{.7335} & .7571 & .7842 \\
($n=3{,}944$) & AR & .3494 & .4635 & .5319 & .5809 & .6174 & .6488 & .6747 & .6978 & .7125 & .7277 & .7497 & .7792 \\
& TIGER & .3474 & .4612 & .5307 & .5834 & .6273 & .6509 & .6795 & .7023 & \textbf{.7208} & \textbf{.7335} & \textbf{.7619} & \textbf{.7896} \\
\bottomrule
\end{tabular}
\caption{Recall at the 12 depths in main-paper Figure~3, evaluated on the
same 6,040 users using paired seed-42 checkpoints and matched Top-200
outputs. This CQG-Single checkpoint has R@10 \(0.3063\), rather than
\(0.3036\) in the primary seven-seed Table~\ref{tab:ml1m-seven-seed}.
CQG-Single--TIGER differences at \(K=10,50,90\) are
\(0.0071,0.0093,0.0040\); all paired-bootstrap 95\% CIs include zero
(10,000 resamples). Bold denotes the best method; ties are bolded.}
\label{tab:crossover}
\end{table}

\subsection{Computing Infrastructure}
\label{app:infrastructure}

Experiments ran in a Linux container with one NVIDIA A800 80GB PCIe GPU
(81,920 MiB device memory), 14 allocated CPU cores, and approximately 120GB
of host-visible memory. The software environment uses Python 3.12.3,
PyTorch 2.8.0+cu128, CUDA 12.8 as bundled with PyTorch, cuDNN 9.10.2,
Transformers 5.13.1, NumPy 2.3.2, scikit-learn 1.9.0, and FAISS-CPU 1.14.3.
Training uses bfloat16 automatic mixed precision where indicated; evaluation
metrics are accumulated outside autocast.

\subsection{Dataset Splits}
\label{app:splits}

All datasets are converted into chronological user sequences. For each user,
we use the last interaction for testing, the penultimate interaction for
validation, and all earlier interactions for training. Users with insufficient
history after filtering are removed. Evaluation is full-catalog unless
otherwise stated: the target item is ranked against all candidate items rather
than against sampled negatives. Cross-dataset accuracy uses
\(K\in\{5,10,20\}\); the ML-1M crossover curve uses the measured depths
\(K\in\{10,20,30,40,50,60,70,80,90,100,120,150\}\).

\subsection{Evaluation Metrics}
\label{app:metrics}

For a user $u$ with held-out target item $i_u^\star$, Recall@$K$ is one if $i_u^\star$ appears in the top-$K$ list and zero otherwise. NDCG@$K$ discounts a hit by the target rank. We distinguish two output-based support metrics. The SID target-support rate is
\[
\mathrm{TargetSupport@B} =
\frac{1}{|\mathcal{U}|}
\sum_{u\in\mathcal{U}}
\mathbf{1}\{i_u^\star\in \mathrm{SIDOutput}_B(u)\},
\]
where $\mathrm{SIDOutput}_B(u)$ is the final ranked list of at most $B$
valid, distinct, history-filtered items produced by constrained SID decoding.
When filtering would leave fewer than $B$ items, the internal beam is expanded
until the list is filled or the valid search space is exhausted. Completed
sequences require an exact catalog-code match; no numeric nearest-code mapping
is used. This is a per-user target-conditioned quantity and is closely related
to Recall@$B$; it must not be interpreted as catalog coverage. For a common
finite-output comparison across methods, we define
\[
\mathrm{TargetInOutput@K}(M)=
\frac{1}{|\mathcal U|}
\sum_{u\in\mathcal U}
\mathbf{1}\{i_u^\star\in\mathrm{TopK}_M(u)\}.
\]
This common finite-output measure is Recall@$K$ under our single-target
leave-one-out protocol. Let
\(\mathcal I_{\mathrm{test}}=\{i_u^\star:u\in\mathcal U\}\) be the set of
distinct held-out target items. We additionally define
\[
\mathrm{TargetItemSupport@K}(M)=
\frac{
\left|\left\{i\in\mathcal I_{\mathrm{test}}:
\exists u,\ i_u^\star=i\ \land\ i\in\mathrm{TopK}_M(u)\right\}\right|
}{
|\mathcal I_{\mathrm{test}}|
}.
\]
Unlike user-weighted Recall, this measures how many distinct target-item types
are successfully retrieved at least once. For any method $M$, empirical union
coverage at a matched returned-list size $K$ is
\[
\mathrm{CatalogCoverage@K}(M)=
\frac{\left|\bigcup_{u\in\mathcal U}\mathrm{TopK}_M(u)\right|}
{|\mathcal I|}.
\]
We report this same output-set quantity for TIGER, CQG-Single, and CQG-AR. Catalog-wide scoreability of a dense interface is a separate property and does not imply 100\% CatalogCoverage@$K$.

Our primary diagnostics directly report item-balanced support in
Table 2 in the main paper, matched union coverage in
Table 4 in the main paper, and explicit training-seed variation in
Table~\ref{tab:ml1m-seven-seed}. These quantities separately address target
imbalance, output diversity, and training-run uncertainty; within-seed item
resampling is not used as a proxy for training-seed variation.

\begin{figure}[!t]
\centering
\includegraphics[width=1\textwidth]{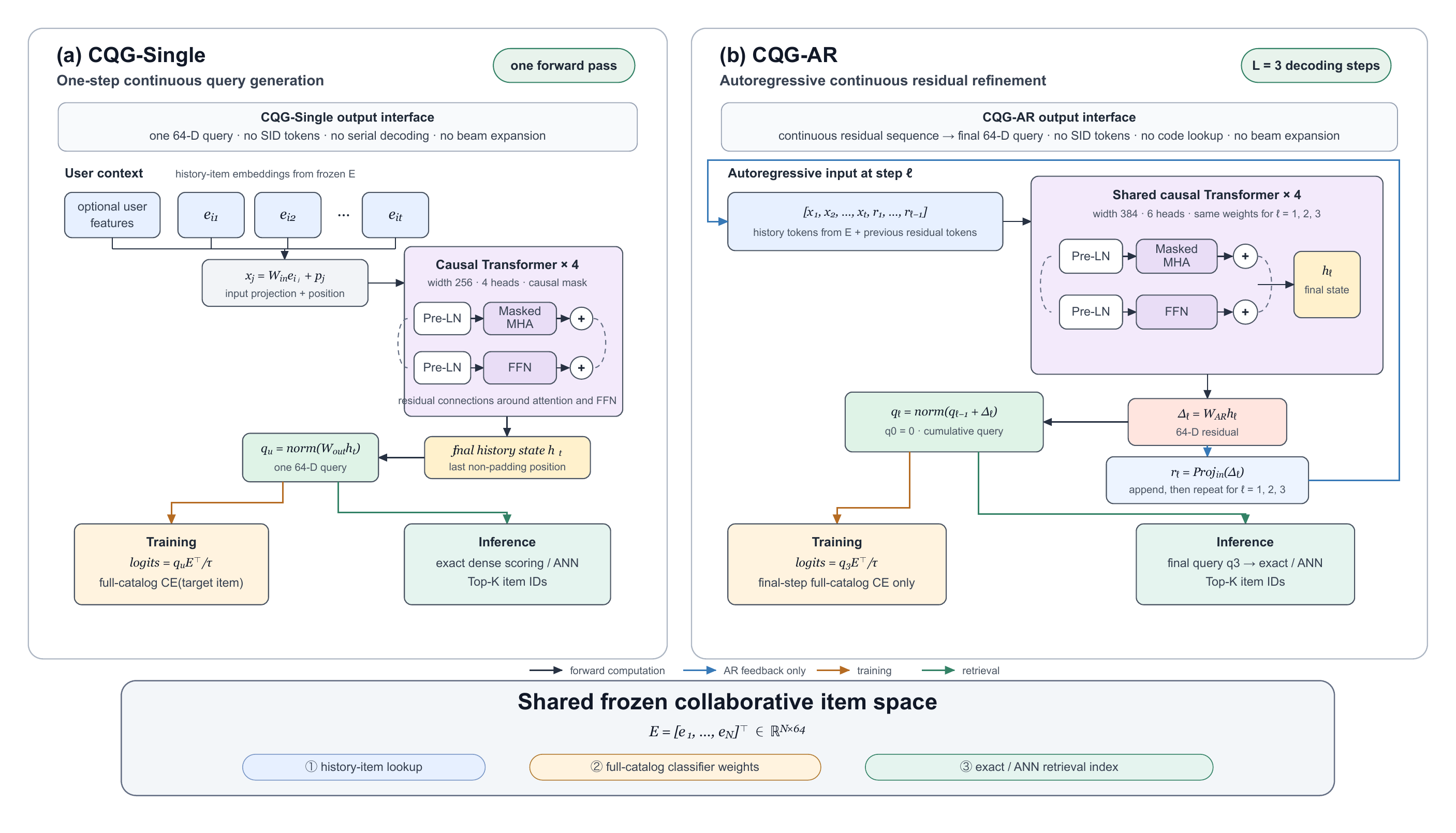}
\caption{Transformer-level architectures of CQG-Single and CQG-AR. Both
methods consume history-item embeddings from the same frozen item table
\(E\). CQG-Single produces one continuous query in a single forward pass,
whereas CQG-AR autoregressively appends projected residual tokens and retrieves
with the final cumulative query. Both use the frozen table as the full-catalog
classifier and exact/ANN retrieval index.}
\label{fig:cqg-architecture}
\end{figure}

\section{Model and Training Details}
\label{app:training}

\subsection{Frozen Collaborative Item Space}
\label{app:item-encoder}

\paragraph{SASRec backbone and examples.}
The item space is learned by a SASRec encoder trained from scratch on the
training portion of each dataset; no text, image, T5, or language-model
features are used in the main item encoder.  The model contains a trainable
item lookup table \(V\in\mathbb{R}^{N\times64}\), learned positional
embeddings, and two causal self-attention blocks with two heads, hidden
dimension 64, and dropout 0.2.  For a chronological training sequence
\((i_1,\ldots,i_T)\), every position \(t=2,\ldots,T\) yields
\[
  ([i_{\max(1,t-L)},\ldots,i_{t-1}],\,i_t),
\]
where \(L=50\) on ML-1M and \(L=20\) on Beauty and Sports.  Histories are
left-padded with item ID 0; the padding embedding is fixed at zero and is
excluded from evaluation.

\paragraph{BPR pretraining and checkpoint selection.}
Let \(h_t\) be the final SASRec hidden state for a history, \(i_t\) its
positive next item, and \(i^-\) one item drawn uniformly from IDs
\(\{1,\ldots,N-1\}\).  SASRec is optimized with
\[
  \mathcal L_{\mathrm{BPR}}
  =-\log\sigma(h_t^\top V_{i_t}-h_t^\top V_{i^-}).
\]
We use Adam with learning rate \(10^{-3}\) and batch size 256.  Every five
epochs, validation ranks the held-out validation item against the full
catalog (masking padding).  We retain the checkpoint with the highest
validation NDCG@10.  The maximum is 200 epochs for ML-1M and Beauty and 400
for Sports; checkpoint selection, rather than the last epoch, determines the
exported representation.

\paragraph{Export, normalization, and sharing.}
After loading the selected checkpoint, we export the lookup parameters
directly,
\[
  E=\texttt{SASRec.item\_emb.weight}\in\mathbb{R}^{N\times64}.
\]
Thus \(e_i\) is a global collaborative item parameter, not a contextual
SASRec hidden state.  We freeze \(E\) for every downstream run.  Its
non-padding rows are L2-normalized for cosine retrieval, CE, BPR, and
InfoNCE; the MSE-trained CQG-Single objective regresses to the raw exported
vector.  The same frozen table is used (i) to embed history IDs,
(ii) as the MSE/InfoNCE target table or the full-catalog CE classifier,
(iii) as the exact or ANN retrieval index, and (iv) as the input to the
hierarchical clustering used by our TIGER-style baseline. This sharing
prevents differences in the item encoder from explaining differences among
the output interfaces. Each dataset uses one fixed, checkpoint-selected
SASRec export for all reported downstream seeds. Seeds 42--48 on ML-1M and
42--44 on Beauty/Sports change downstream initialization, data order, and
optimization; TIGER additionally reruns hierarchical clustering for each
indexed seed. The ML-1M table has shape \(3417\times64\).

\subsection{CQG-Single Architecture and Optimization}
\label{app:cqg-single-training}

CQG-Single has no pretrained backbone and does not update \(E\).  Frozen
64-dimensional history vectors are projected to width 256 and combined with
learned positional embeddings.  Four pre-norm causal Transformer blocks,
each with four attention heads, feed-forward dimension 1024, and dropout 0.1,
produce the final valid history state.  A linear head maps this state back to
64 dimensions and the query is L2-normalized before scoring the normalized
catalog.  The CE configuration uses Adam, learning rate \(10^{-3}\), batch
size 256, at most 50 epochs, validation every five epochs, and early stopping
with patience 15 based on validation NDCG@10.  MSE and InfoNCE change only
the objective and their stated learning-rate/epoch settings in
Table~\ref{tab:hparams}; the history protocol and frozen target table remain
the same.

\subsection{CQG-AR Architecture and Optimization}
\label{app:cqg-ar-training}

CQG-AR is also trained from scratch with frozen \(E\).  It projects history
vectors from 64 dimensions to width 384 and uses four pre-norm causal
Transformer blocks with six heads, feed-forward dimension 1024, and dropout
0.1.  A learned BOS vector starts continuous decoding.  At step \(\ell\), a
linear head followed by \(\tanh\) produces
\(\Delta_\ell\in\mathbb{R}^{64}\); a learned scalar controls its magnitude,
the cumulative query is initialized as \(q^{(0)}=\mathbf{0}\), and
\[
 q^{(\ell)}=\operatorname{norm}(q^{(\ell-1)}+\Delta_\ell).
\]
The residual is projected back to width 384, tagged with a generated-token
type embedding and a position embedding, and appended as the next causal
decoder input.  The main configuration uses three steps, so the serial
outputs are \((\Delta_1,\Delta_2,\Delta_3)\), but all three refer to the same
next-item target \(i_t\); they are not labels for three SID levels.

For the final CE recipe, only \(q^{(3)}\) receives direct catalog CE
supervision (\texttt{intermediate\_weight}=0); earlier steps receive gradient
through their effect on later decoding.  The depth ablation instead uses
intermediate weight 0.3 for the non-final queries, always with the same
next-item label.  CQG-AR uses AdamW with learning rate \(3\times10^{-4}\),
weight decay 0.01, batch size 1024, at most 30 epochs, bfloat16 automatic
mixed precision, gradient-norm clipping at 1.0, and early stopping with
patience 5 on validation NDCG@10.  The selected checkpoint is evaluated
against the full catalog after appending the validation item to the test
history and masking all previously observed items.

\subsection{Hyperparameters}
\label{app:hparams}

\begin{table}[H]
\centering
\small
\setlength{\tabcolsep}{5pt}
\begin{tabular}{lrrrr}
\toprule
Method & Item $d$ & Hidden & Layers & Heads \\
\midrule
SASRec item encoder & 64 & 64 & 2 & 2 \\
CQG-Single MSE & 64 & 256 & 4 & 4 \\
CQG-Single InfoNCE & 64 & 256 & 4 & 4 \\
CQG-Single CE & 64 & 256 & 4 & 4 \\
CQG-AR & 64 & 384 & 4 & 6 \\
CQG-AR sampled CE & 64 & 384 & 4 & 6 \\
Qwen2.5-7B hybrid probe & 64 & 3584 & 28 & 28 \\
TIGER-style SID decoder & 64 & 384 & 4 & 6 \\
\bottomrule
\end{tabular}

\vspace{4pt}

\begin{tabular}{lrrrl}
\toprule
Method & LR & Batch & Epochs/Steps & Objective \\
\midrule
SASRec item encoder & 1e-3 & 256 & 200/400 max. & BPR, one uniform negative \\
CQG-Single MSE & 1e-3 & 256 & 50 max. & MSE to target embedding \\
CQG-Single InfoNCE & 1e-4 & 256 & 100 max. & In-batch InfoNCE, $\tau=0.07$ \\
CQG-Single CE & 1e-3 & 256 & 50 max. & Full-catalog softmax CE \\
CQG-AR & 3e-4 & 1024 & 30 max. & AR residuals + full-catalog CE \\
CQG-AR sampled CE & 3e-4 & 1024 & 30 max. & AR residuals + 256 negatives \\
Qwen2.5-7B hybrid probe & 5e-5 & 32$\times$8 & 10 & InfoNCE with projection head \\
TIGER-style SID decoder & 1e-3 & 256 & 50K steps & Autoregressive CE over SID tokens \\
\bottomrule
\end{tabular}
\caption{Training hyperparameters for the three datasets. ``Hidden'' is the
Transformer width; all continuous queries remain 64-dimensional. The
Qwen2.5-7B probe uses its 3,584-dimensional, 28-layer, 28-attention-head
backbone and a 64-dimensional projection head. SASRec uses at most 200 epochs on
ML-1M/Beauty and 400 on Sports. Sequence length is 50 for ML-1M and 20 for
Beauty/Sports; ``max.'' denotes early-stopped training.}
\label{tab:hparams}
\end{table}

\subsection{Continuous-Query Dataloader}
\label{app:cqg_dataloader}

CQG-Single and CQG-AR use the same next-item examples derived from user histories. Each training example contains a padded history sequence and the next-item target. The history item IDs are mapped to frozen item embeddings before entering the query generator; the target ID is used either to index the target embedding for MSE/InfoNCE or to index the full-catalog label for CE. A batch therefore has the form
\[
(\mathbf{H}, \mathbf{y}) =
\{([i_{t-L},\ldots,i_{t-1}], i_t)\}_{b=1}^{B},
\]
where $L$ is the maximum sequence length and $B$ is the batch size. Padding positions are masked and never treated as candidate items. This dataloader is intentionally the same sequential next-item protocol used by SASRec, so the comparison isolates the output interface rather than changing the recommendation task.

For example, one actual ML-1M training pair is
\[
([2758,1069,1445,862,1979],\,1520).
\]
The model input is the \(5\times64\) matrix obtained by looking up the five
history IDs in frozen \(E\). Under CE, the label is the integer 1520 among
3,416 non-padding catalog items; the implementation stores these items in a
3,417-row table whose row 0 is masked padding. Under MSE or InfoNCE, the
corresponding target is row \(E_{1520}\). At test time the validation item is appended to the
training history, the final item is the held-out target, and all items already
present in the resulting history are masked from ranking.

\subsection{Continuous Query Training Objectives}
\label{app:cqg_losses}

The MSE variant regresses the generated query $q_u$ to the frozen target embedding $e_{i^\star}$ (the MSE-trained CQG-Single run uses the raw SASRec export):
\[
\mathcal{L}_{\mathrm{MSE}} =
\|q_u - e_{i^\star}\|_2^2.
\]
The InfoNCE variant treats other targets in the batch as negatives:
\[
\mathcal{L}_{\mathrm{NCE}} =
-\log
\frac{\exp(q_u^\top e_{i^\star}/\tau)}
{\sum_{j\in\mathcal{B}}\exp(q_u^\top e_j/\tau)}.
\]
The CE variant scores the full frozen catalog:
\[
\mathcal{L}_{\mathrm{CE}} =
-\log
\frac{\exp(q_u^\top e_{i^\star}/\tau)}
{\sum_{j=1}^{N}\exp(q_u^\top e_j/\tau)}.
\]
For CQG-AR, \(q_u=q^{(3)}\) in the final recipe.  All autoregressive steps
share the same next-item target; there is no separate label per residual.
All variants can score every encoded item by dense similarity at inference
time. Their empirical CatalogCoverage@$K$ is nevertheless measured from the
union of finite top-$K$ outputs.

\section{SID Construction and Decoder Training}
\label{app:sid}

\subsection{Semantic-ID Construction}
\label{app:sid_construction}

Our TIGER-style baseline hierarchically clusters the frozen SASRec item table
into depth-3 paths. Collisions are resolved before decoder training by
appending deterministic residual identifiers ordered by item ID. On ML-1M,
99.4\% of items are already unique after the three cluster levels; the
remaining items become one-to-one catalog codes after this tie-break.

\subsection{TIGER T5/RQ-VAE Reproduction Recipe}
\label{app:tiger_recipe}

The Beauty T5/RQ-VAE run uses fixed codes and three decoder seeds; the
tokenizer is not retrained.
Across these runs, 99.48\% of generated sequences are valid SIDs, while the
union of Top-100 outputs covers \(44.0\%\pm13.7\%\) of the catalog, rising
from \(14.0\%\pm7.6\%\) in Q1 to \(88.7\%\pm13.0\%\) in Q4.

Its R@10 is \(0.0408\pm0.0016\), below the reference implementation. Thus it
confirms popularity-skewed SID coverage under a second construction, but not
published TIGER accuracy.

\subsection{Valid-Prefix Decoding}
\label{app:tiger_dataloader}

The SID decoder uses the same history--target examples as the continuous
models but predicts code tokens. At inference, a catalog trie masks invalid
prefixes, completed codes use exact item lookup, historical items are removed,
and the internal beam expands when necessary to fill the requested number of
valid, distinct items. Checkpoint selection uses this same constrained
protocol.

\section{Additional Ablations}
\label{app:ablations}

\subsection{Objective, Catalog-Visibility, and Temperature Diagnostics}
\label{app:loss-objectives}

\paragraph{Objective and catalog visibility.}
\iffalse
\begin{table}[H]
\centering
\small
\begin{tabular}{lrrr}
\toprule
Objective & R@10 & NDCG@10 & R@100 \\
\midrule
Full-catalog CE & \textbf{0.2982 \(\pm\) 0.0042} & \textbf{0.1727 \(\pm\) 0.0020} & \textbf{0.6631 \(\pm\) 0.0048} \\
Sampled CE (256 negatives) & 0.2895 \(\pm\) 0.0077 & 0.1663 \(\pm\) 0.0034 & 0.6592 \(\pm\) 0.0104 \\
\bottomrule
\end{tabular}
\caption{ML-1M objective-visibility control (mean $\pm$ sample SD; seeds 42--44). Reducing catalog visibility to 256 sampled negatives lowers R@10 by 0.0087 but retains competitive accuracy, so objective visibility explains part, not all, of the continuous model's behavior.}
\label{tab:objective-control}
\end{table}
\fi

Full-catalog CE yields the highest Recall@10 in the frozen item space.
Replacing it with 256 sampled negatives lowers CQG-AR mean R@10 from
\(0.2982\) to \(0.2895\), while remaining above MSE and InfoNCE.
CE-joint is separate because it also changes the item representations.

\iffalse
\begin{table}[H]
\centering
\small
\begin{tabular}{lrr}
\toprule
Objective & Test R@10 & vs. SASRec \\
\midrule
SASRec BPR & 0.2797 & 100.0\% \\
CQG-Single full-catalog CE & \textbf{0.3114} & 111.3\% \\
CQG-Single MSE & 0.2425 & 86.7\% \\
CQG-Single InfoNCE & 0.2386 & 85.3\% \\
CQG-AR full-catalog CE & 0.2982 & 106.6\% \\
CQG-AR MSE & 0.2424 & 86.7\% \\
CQG-AR InfoNCE & 0.2324 & 83.1\% \\
\bottomrule
\end{tabular}
\caption{Frozen-item objective landscape on ML-1M. CQG-Single CE/MSE and all
CQG-AR rows are three-seed means under the corrected test-history protocol;
CQG-Single InfoNCE is an auxiliary diagnostic.}
\label{tab:loss}
\end{table}
\fi

\paragraph{Temperature.}
\label{app:temperature}

\iffalse
\begin{table}[!htbp]
\centering
\small
\begin{tabular}{lr}
\toprule
Objective & Test R@10 \\
\midrule
CQG-Single CE, $\tau=0.07$ & 0.3114 \\
CQG-Single CE, $\tau=0.10$ & 0.3070 \\
CQG-Single CE, $\tau=0.20$ & 0.2846 \\
CQG-Single MSE & 0.2425 \\
CQG-AR CE, $\tau=0.07$ & 0.2972 \\
CQG-AR CE, $\tau=0.10$ & 0.2886 \\
CQG-AR CE, $\tau=0.20$ & 0.2725 \\
\bottomrule
\end{tabular}
\caption{Temperature ablation on ML-1M under the corrected test-history
protocol. CQG-Single values are three-seed means; CQG-AR values are aligned
seed-42 controls. Lower temperatures improve shallow retrieval in the tested
range.}
\label{tab:temperature}
\end{table}
\fi

Lower temperatures monotonically improve shallow retrieval in the tested
range. CQG-Single R@10 decreases from \(0.3114\) at
\(\tau=0.07\) to \(0.3070\) at \(0.10\) and \(0.2846\) at \(0.20\);
the aligned seed-42 CQG-AR control decreases from \(0.2972\) to \(0.2886\)
and \(0.2725\). The MSE-trained CQG-Single reference reaches \(0.2425\).

\begin{figure}[H]
\centering
\includegraphics[width=\textwidth]{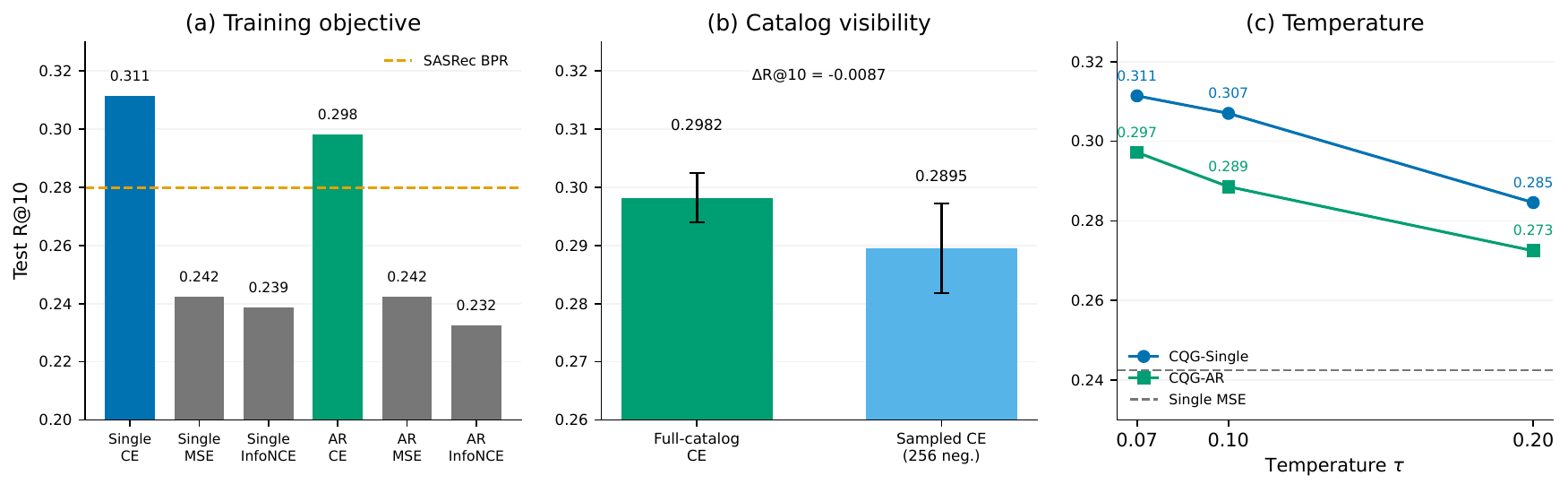}
\caption{Unified ML-1M objective diagnostics. (a) Full-catalog CE gives the
highest R@10 for both continuous decoders in the shared frozen item space.
(b) Replacing full-catalog CE with 256 sampled negatives reduces CQG-AR mean
R@10 by 0.0087. Error bars denote sample SD over seeds 42--44.
(c) Lower softmax temperatures improve R@10 for both continuous decoders in
the evaluated range; the dashed line is the CQG-Single MSE reference.}
\label{fig:ce-mse}
\end{figure}

\section{LLM Backbone Diagnostic}
\label{app:llm}

\begin{figure}[!htbp]
\centering
\includegraphics[width=0.45\textwidth]{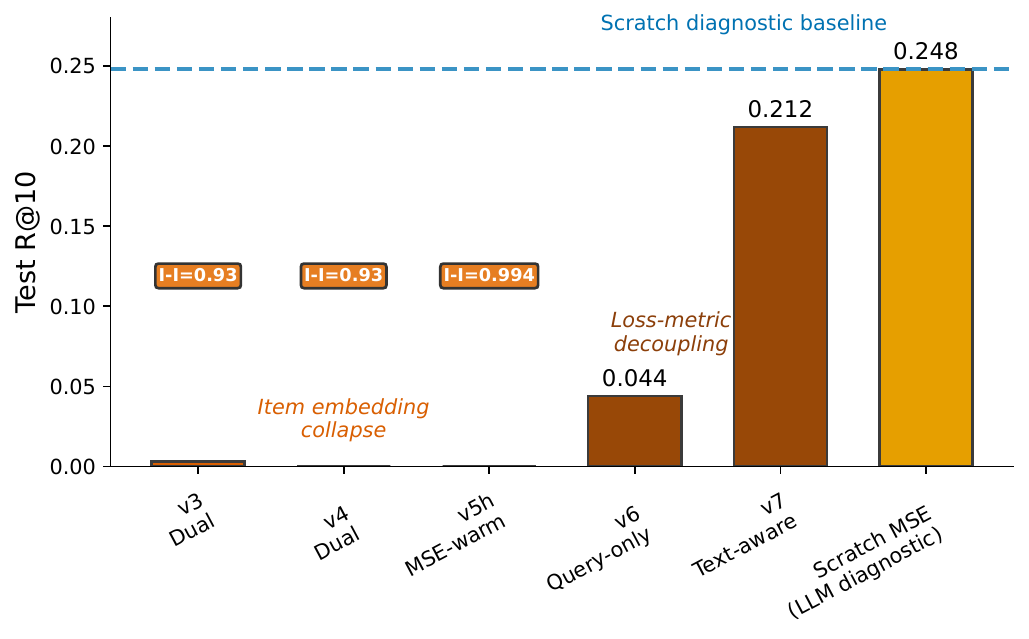}
\caption{LLM backbone diagnostics. Pretrained language backbones do not
automatically align with the frozen collaborative item space; the best
text-aware Qwen hybrid still trails the single-run, MSE-frozen scratch query
Transformer used in this diagnostic (R@10 = 0.248). This checkpoint is
separate from the three-seed CQG-Single MSE result in
Figure~\ref{fig:ce-mse}. In the figure, v3/v4 are jointly trained dual-tower
variants, v5h is an MSE-warm-started dual tower, v6 trains only the query side
against a frozen item table, and v7 is the frozen-item text-aware hybrid.
The first three collapse the item space; v6/v7 instead exhibit loss--metric
decoupling. These are single-run diagnostics, not primary baselines.}
\label{fig:llm_diagnostic}
\end{figure}

No tested Qwen configuration outperforms the scratch Transformer
(Figure~\ref{fig:llm_diagnostic}). Joint training collapses the item space
(item--item cosine 0.9936); freezing it avoids collapse but leaves Recall@10
below the SASRec gate. The best hybrid reaches R@10 = 0.2118 versus 0.248 for
the single-run scratch model and trails it at every cutoff. This scratch
checkpoint is distinct from the three-seed MSE result in
Figure~\ref{fig:ce-mse}.

\begin{table}[!htbp]
\centering
\small
\begin{tabular}{lrrrrrr}
\toprule
Model & R@5 & R@10 & R@20 & R@50 & R@70 & R@90 \\
\midrule
\shortstack[l]{Scratch Transformer\\(LLM diagnostic)} & \textbf{0.172} & \textbf{0.248} & \textbf{0.345} & \textbf{0.498} & \textbf{0.555} & \textbf{0.595} \\
Qwen text-aware hybrid & 0.134 & 0.212 & 0.308 & 0.460 & 0.517 & 0.564 \\
\bottomrule
\end{tabular}
\caption{Single-run recall curve comparison between the MSE-frozen scratch
Transformer used in the LLM diagnostic and the best tested LLM hybrid
configuration. The scratch row is not the three-seed CQG-Single MSE aggregate
reported in Figure~\ref{fig:ce-mse}. Bold denotes the best value in each
column.}
\label{tab:llmcurve}
\end{table}

% The full recall curve confirms the diagnostic: the best tested LLM hybrid
% underperforms the scratch Transformer at every reported depth.

\FloatBarrier
\bibliographystyle{plainnat}
\bibliography{references}

\end{document}